\documentclass{article}

\PassOptionsToPackage{numbers}{natbib}
\usepackage[preprint]{neurips_2024}

\usepackage[utf8]{inputenc} % allow utf-8 input
\usepackage[T1]{fontenc}    % use 8-bit T1 fonts
\usepackage{hyperref}       % hyperlinks
\usepackage{url}            % simple URL typesetting
\usepackage{booktabs}       % professional-quality tables
\usepackage{amsfonts}       % blackboard math symbols
\usepackage{nicefrac}       % compact symbols for 1/2, etc.
\usepackage{microtype}      % microtypography
\usepackage{xcolor}         % colors

\usepackage[newfloat, frozencache=true,cachedir=_minted_output]{minted} % To use cache

\usepackage{graphicx}
\usepackage{caption}
\usepackage{subcaption}
\usepackage{mathtools}

\usepackage[ruled,vlined]{algorithm2e}
\usepackage{tcolorbox}
\tcbuselibrary{minted,breakable,xparse,skins}

\definecolor{bg}{gray}{0.95}
\DeclareTCBListing{mintedbox}{O{}mO{}}{%
	breakable=true,
	listing engine=minted,
	listing only,
	minted language=#2,
	minted style=default,
	minted options={%
		linenos,
		gobble=0,
		breaklines=true,
		breakafter=,,
		fontsize=\small,
		escapeinside=||,
		numbersep=8pt,
		tabsize=2,
		encoding=utf8,
		#1},
	boxsep=0pt,
	left skip=0pt,
	right skip=0pt,
	left=25pt,
	right=0pt,
	top=3pt,
	bottom=3pt,
	arc=5pt,
	leftrule=0pt,
	rightrule=0pt,
	bottomrule=2pt,
	toprule=2pt,
	colback=bg,
	colframe=orange!70,
	enhanced,
	overlay={%
		\begin{tcbclipinterior}
			\fill[orange!20!white] (frame.south west) rectangle ([xshift=20pt]frame.north west);
	\end{tcbclipinterior}},
	#3}

\newtheorem{example}{Example}
\newtheorem{definition}{Definition}[section]

\newcommand{\reels}{\mathbb{R}}
\newcommand{\complex}{\mathbb{C}}
\newcommand{\normltwo}{L^2}
\newcommand*\diff{\mathop{}\!\mathrm{d}}
\newcommand{\norm}[1]{\left\lVert#1\right\rVert}

\title{Differentiable programming across the PDE and Machine Learning barrier}

\author{
    Nacime Bouziani\\
    I-X Centre for AI In Science,\\
    Imperial College London, UK \\
  \texttt{n.bouziani18@imperial.ac.uk} \\
  \And
  David A. Ham\\
  Department of Mathematics, \\
  Imperial College London, London, UK \\
  \texttt{david.ham@imperial.ac.uk} \\
  \And
  Ado Farsi \\
  Department of Earth Science and Engineering, \\
  Imperial College London, London, UK \\
  \texttt{ado.farsi@imperial.ac.uk}\\
}

\begin{document}

\maketitle

\begin{abstract}
  The combination of machine learning and physical laws has shown immense potential for solving scientific problems driven by partial differential equations (PDEs) with the promise of fast inference, zero-shot generalisation, and the ability to discover new physics. Examples include the use of fundamental physical laws as inductive bias to machine learning algorithms, also referred to as physics-driven machine learning, and the application of machine learning to represent features not represented in the differential equations such as closures for unresolved spatiotemporal scales. However, the simulation of complex physical systems by coupling advanced numerics for PDEs with state-of-the-art machine learning demands the composition of specialist PDE solving frameworks with industry-standard machine learning tools. Hand-rolling either the PDE solver or the neural net will not cut it. In this work, we introduce a generic differentiable programming abstraction that provides scientists and engineers with a highly productive way of specifying end-to-end differentiable models coupling machine learning and PDE-based components, while relying on code generation for high performance. Our interface automates the coupling of arbitrary PDE-based systems and machine learning models and unlocks new applications that could not hitherto be tackled, while only requiring trivial changes to existing code. Our framework has been adopted in the Firedrake finite-element library and supports the PyTorch and JAX ecosystems, as well as downstream libraries.
\end{abstract}

\section{Introduction}
\label{sec:introduction}

Partial differential equations (PDEs) are central to describing and modelling complex physical systems that arise in many disciplines across science and engineering. This modelling is most effective when physical systems closely follow the, frequently idealised, assumptions used to derive the PDE. The models of real phenomena of interest to scientists and engineers are, regrettably, seldom so straightforward. Alongside the fundamental physical laws expressed as PDEs are various empirical parametrisations, closures, and regularisation terms which represent aspects of the system for which a more fundamental model is either not known or is practically infeasible for some reason. In addition, traditional PDE solvers are, in most cases, notoriously expensive, and do not necessarily scale with data observations.

To combat these limitations, the combination of machine learning and PDEs has been proposed across different disciplines, ranging from geoscience \cite{L_Zanna_2019, shi_deep_2020, zhang_chapter_2022} to structural mechanics \cite{haghighat_physics-informed_2021, lai_structural_2021} to name but two fields. A first approach consists in embedding PDE-based components into machine learning algorithms, such as solving a PDE to guide training and/or compute coarse-resolution features \cite{um2020sol, belbute-peres_combining_2020, nguyen_tnet_2022}, assembling the residual of a PDE \cite{raissi_physics-informed_2019, kharazmi_variational_2019, karniadakis_physics-informed_2021}, or encoding divergence-free conditions for incompressible fluid simulations \cite{richter-powell_neural_2022}.

Machine learning models can also be embedded in PDE-based systems. Examples include the use of deep learning-based regularisers for inverse problems \cite{zhang_chapter_2022, shi_deep_2020}, or incorporating ML-based constitutive relations, closures, and parametrisations in PDE systems \cite{L_Zanna_2019, crilly2024learning}. Notably, this approach also entails the training of machine learning architectures embedded in differential equations, also referred to as universal differential equations \cite{rackauckas_universal_2021}.

The composition of advanced numerics for PDEs with state-of-the-art machine learning requires a vast range of capabilities spanning several disciplines that are currently not at expert level in a single framework and are only available in PDE-specific or ML-specific software. Examples of capabilities on the PDE side include: mesh generation for complex geometries, sophisticated spatial and temporal discretisations, boundary conditions, state-of-the-art solvers, adjoint capabilities, coupled physics, and advanced parallelism. Examples of capabilities on the ML side include: advanced model architectures, data processing (e.g. data loading), training optimisers, algorithmic differentiation, data and model parallelism, and inference optimisation techniques.

Instead of hand-rolling the PDE part in a machine learning framework or vice-versa, we present a coupling of differentiable programming tools which embody the mathematical formulation of each part of the system. Our interface allows the coupling of PDE-based systems implemented in the Firedrake framework \cite{FiredrakeUserManual} with ML models implemented in machine learning tools such as PyTorch \cite{PyTorch} or JAX \cite{jax2018github}. This enables researchers, engineers, and domain experts to design hybrid models that seamlessly compose with the framework pipeline they are defined in, thereby allowing the training of ML algorithms with differentiable physical constraints and/or the solution of PDE-based systems with ML components. The resulting software environment is embedded in Python and ensures easy and efficient interoperability, which allows end users to leverage the rich ecosystem of Python libraries.

\section{Related Work}
\label{sec:related_work}

Most of the existing works focus on incorporating PDE systems into machine learning frameworks, thereby yielding differentiable physics constraints that can be incorporated into ML algorithms thanks to the algorithmic differentiation pipelines of these frameworks. Some frameworks have specialised in particular applications such as \textit{XLB} \cite{ataei_xlb_2024} and \textit{PhiFlow} \cite{holl_learning_2020} for fluid simulations, or \textit{Adept} \cite{joglekar_machine_2023} for plasma physics. Others have focused on implementing particular numerical solvers such as \textit{JAX-FEM} \cite{xue_jax-fem_2023} for the finite element method. However, these approaches only implement certain PDE-based capabilities (solvers, discretisations, boundary conditions treatment) bound to specific use cases.

In \cite{rackauckas_universal_2021}, the authors proposed \textit{SciML}, a Julia ecosystem for scientific machine learning. Their framework interfaces with the FEniCS finite element framework \cite{Fenics_2012}, which is comparable to Firedrake, by wrapping some of its capabilities. However, they lack the adjoint capabilities needed for differentiating through PDE-based constraints. In addition, their interface does not allow to embed ML models into arbitrary PDE systems. Finally, \cite{belbute-peres_combining_2020}, embeds \textit{SU2} \cite{economon_su2_2016} in PyTorch. This approach is conceptually similar to ours. However, it is limited to fluid simulations, and only allows for embedding PDE solvers into PyTorch with differentiation capabilities for a restricted set of physical quantities only. 

To the best of our knowledge, our framework is the first capable of embedding completely general PDE systems into arbitrary ML models and vice-versa, with state-of-the-art capabilities, and in an end-to-end differentiable fashion.

\section{Differentiable programming}
\label{sec:diff_programming}

To couple PDEs and ML, we need to be able to couple the evaluation of both components but also to couple their differentiation as gradient calculation is critical in both machine learning and PDE-based modelling. Examples include the use of backpropagation for training neural networks, the use of Newton-type methods for solving PDEs, or the use of adjoint methods for solving inverse problems driven by PDEs. This need for differentiation can be addressed by employing differentiable programming.

\begin{definition}
    \label{def:diff_programming}
    Differentiable programming is a programming paradigm in which programs comprise differentiable operators between finite-dimensional Hilbert spaces composed together using differentiable high order functions.
\end{definition}

Differentiable programming allows computer programs to be seemlessly differentiated using algorithmic differentiation (AD), enabling gradient-based optimisation of parameters in the program. The differentiable programming paradigm is universal in machine learning but in the field of PDE software the differentiable programming paradigm (as opposed to lower-level, less automated, AD approaches) has been less widely adopted. The only general PDE frameworks embodying the approach are  Firedrake \cite{FiredrakeUserManual} and pre-2019 versions of FEniCS \cite{Fenics_2012}.

Given differentiable programming tools for both ML and PDEs, the key contribution of this work lies in defining the differentiable composition operations between the two.

\subsection{Tangent linear and adjoint models on Hilbert spaces}

Readers familiar with ML will be familiar with the two core operations of  algorithmic differentiation: tangent-linear mode, also referred to as forward mode, and the adjoint mode, also referred to as reverse mode, of which backpropagation is a special case. In contrast with ML where the control parameters are typically in $\reels^n$, PDE simulations, and in particular the finite element method, typically involve finite-dimensional function spaces that are discretisations of infinite-dimensional Hilbert spaces. The key distinction is that $\reels^n$ is self dual: the difference between tensors and linear functionals mapping tensors to $\reels$ is merely the difference between a row and column vector. In contrast, the primal and dual spaces for more general Hilbert spaces are distinct. This produces more general derivative definitions:

\begin{definition}
\label{def:tlm_adjoint}
Let $U$ and $V$ be Hilbert spaces, and $f : U \rightarrow V$ be a Fr\'echet differentiable function. Let $u \in U$, the \textbf{tangent linear model} of $f$ is the linear form  $\mathcal{J}_{f, u}: U \rightarrow V$ defined by
\begin{equation}
    \label{eq:definition_TLM}
    \mathcal{J}_{f, u} (v) \vcentcolon = \frac{\diff f(u; v)}{\diff u} \quad \forall v \in U
\end{equation}
where $\frac{df(u; v)}{du}$ is the G\^ateaux derivative of $f$ with respect to $u$ in the direction $v$. Likewise, the \textbf{adjoint model} of $f$ is the linear form  $\mathcal{J}_{f}^{*}: V^{*} \rightarrow U^{*}$ defined by
\begin{equation}
    \label{eq:definition_Adjoint}
    \mathcal{J}_{f, u}^{*} (v) \vcentcolon = \frac{\diff f(u; v)}{\diff u} \quad \forall v \in V^{*}
\end{equation}
where $\frac{\diff f(u; v)}{\diff u}$, for $v \in V^{*}$, is the adjoint of the G\^ateaux derivative of $f$ with respect to $u$ in the direction $v$, and where $U^{*}$ and $V^{*}$ are the dual spaces of $U$ and $V$, i.e. the spaces of all bounded linear functionals on $U$ and $V$, respectively.
\end{definition}

For example, let $V$ be a Hilbert space and let $u$ be the solution of a PDE defined by:
\begin{equation}
\label{eq:variational_form}
F(u, m; v) = 0\quad \forall v \in V    
\end{equation}
where $F$ is the variational form of the PDE, and $m$ is a known parameter. Then, backpropagating through the solution $u$ is equivalent to computing the adjoint model of $u$, which can be written as:
\begin{equation}
    \label{eq:TLM_PDE}
    \mathcal{J}^{*}_{u, m}(w) = - \frac{\partial F}{\partial m}^{*} \lambda
\end{equation}
for all $w \in V^{*}$ and where $\lambda \in V$ is the solution of the adjoint equation defined as:
\begin{equation}
    \label{eq:Adjoint_equation}
    \frac{\partial F}{\partial u}^{*} \lambda = w
\end{equation}

See appendix \ref{sec:adjoint_method} for more details. In contrast to ML frameworks, Firedrake supports primal and dual finite element spaces as first class types.

\subsection{Firedrake}

Firedrake \citep{FiredrakeUserManual} is an automated finite element system embedded in Python for the solution of partial differential equations. Firedrake uses the Unified Form Language (UFL) \citep{UFL_2014}, a domain-specific language to provide high-level  differentiable representations of finite element problems. The Firedrake system translates the symbolic specification of variational forms of PDEs expressed in UFL into low-level code for assembling the sparse matrices and vectors of the corresponding finite element problem. The dolfin-adjoint package \citep{farrell_automated_2013, dolfin-adjoint_2019} enables the taping of the composition of UFL operations, and completes the differentiable programming capabilities of Firedrake. In addition, Firedrake also has ensemble parallelism capabilities, which naturally allow batching of operations such as solving a PDE or assembling a variational form when it is embedded in an ML framework.

\section{Differentiable coupling of Firedrake and ML frameworks}
\label{sec:dp_interface}

Our approach relies on the simple key idea that any simulation combining both components can be represented as a single computational graph (DAG). Our contribution is the ability to embed the DAG associated with a Firedrake program into the DAG associated with the program of a given ML framework and vice-versa. The computational graph evaluation and differentiation is delegated to the AD engine of the ML framework for the framework-related nodes, and to \emph{dolfin-adjoint} for Firedrake-based nodes, where both environments rely on code generation ensuring high-performance evaluation of the nodes as well their tangent linear and adjoint models.

\begin{figure}[ht!]
\centering
\includegraphics[width=.70\linewidth]{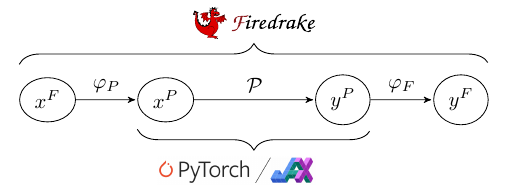}
\caption{Subgraph of the Firedrake computational graph containing PyTorch/JAX operations of interest represented by $\mathcal{P}$, where $P$ refers to PyTorch/JAX variables and $F$ to Firedrake variables. $\varphi_{F}$ and $\varphi_{P}$ represent the casting of a PyTorch/JAX tensor to a Firedrake Function and vice versa.}
\label{fig:coupling_pytorch_firedrake}
\end{figure}

This simple yet powerful high-level coupling, illustrated in figures~\ref{fig:coupling_pytorch_firedrake} and \ref{fig:coupling_firedrake_pytorch}, results in a composable environment that benefits from the full armoury of advanced features and AD capabilities both frameworks offer whilst maintaining separation of concerns.

\subsection{Embedding ML models into PDE systems}

UFL can represent the weak form of essentially any partial differential equation. Here we introduce an expressive, flexible and differentiable interface for incorporating arbitrary operators in UFL, and providing their implementation to Firedrake. This is used to define symbolic machine learning operator objects at the UFL level. 

In our abstraction, a machine learning operator $N$ is defined as an operator mapping $k$ operands to a function space $\mathcal{X}$:
\begin{equation}
\label{def_external_operator_OPERATOR}
\begin{aligned}
N \vcentcolon  W_1 \times \cdots \times W_k &  \longrightarrow \mathcal{X} \\
 u_1,\ \ldots,\ u_k  \quad &\longmapsto N(u_1, \ldots, u_k)
\end{aligned}
\end{equation}
where $(W_{i})_{1\le i\le k}$ and $\mathcal{X}$ are either finite element spaces or $\reels^n$, where the operands $(u_{i})_{1 \le i \le k}$ refer to the inputs of the embedded machine learning operator. 
We refer to appendix \ref{sec:ml_operator_firedrake_ufl} for more details on the symbolic and numerical representation of machine learning operators in Firedrake and UFL.

The evaluation of an ML operator in Firedrake is achieved by calling the machine learning framework considered. For example, solving a PDE containing an ML operator $N(u)$, where $u$ is the model input but also the solution of the PDE, requires the derivative of $N$ with respect to $u$ when using Newton-type methods. This can be achieved by symbolically differentiating $N$ and assembling (evaluating) it. Our interface delegates this derivative evaluation to the ML framework.\\
\begin{listing}[h!]
\captionof{listing}{Outline of the Firedrake ML operator interface showing ML operator evaluation, tangent linear and adjoint models (lines 11, 15, 17). The ML backend is chosen using the module import, e.g. lines 3 can be replaced by line 2 to embed a model defined in JAX instead of PyTorch.}
\label{code:neural_net_interface}
\begin{mintedbox}[highlightlines={8}]{python}
from firedrake import *
# from firedrake.ml.jax import *
from firedrake.ml.pytorch import *
...
# Define PyTorch/JAX model |$\mathcal{P}$|
P = ...
# Define the Firedrake operator wrapping the ML model
N = |\textcolor{blue}{ml\_operator}|(P, function_space=...)

# Evaluate |$N$|
y_F = |\textcolor{blue}{assemble}|(N(x_F))

dNdu = |\textcolor{blue}{derivative}|(N(x_F), x_F)
# Compute |$\mathcal{J}_{N, u}(\delta u)$|
tlm_value = |\textcolor{blue}{assemble}|(|\textcolor{blue}{action}|(dNdu, |$\delta u$|))
# Compute |$\mathcal{J}^{*}_{N, u}(\delta N)$|
adj_value = |\textcolor{blue}{assemble}|(|\textcolor{blue}{action}|(|\textcolor{blue}{adjoint}|(dNdu), |$\delta N$|))
\end{mintedbox}
\end{listing}
Likewise, our interface also seamlessly unlocks complex couplings such as the training of a machine learning model $N(u)$ embedded in a PDE, and where the loss function depends on the PDE solution $u$. This entails computing the gradient of the loss with respect to the model parameters $\theta$. This in turn requires taking the adjoint of $\frac{\partial F}{\partial \theta}$, which by chain rule necessitates taking the adjoint of $\frac{\partial N}{\partial \theta}$.  At the Firedrake level, this can be simply achieved by line 17 in listing \ref{code:neural_net_interface}.

\subsection{Embedding PDE systems into ML frameworks}

Embedding differentiable physical constraints implemented in Firedrake into PyTorch or JAX is facilitated by the fact that the mathematical abstraction surrounding operations in these frameworks is simpler than the one pertaining to finite element problems. Our contribution is twofold: first, we extend dolfin-adjoint to allow for algorithmic differentiation of non-scalar-valued operations. This is required because Firedrake may no longer be responsible for the final reduction to a scalar functional. Second, we introduce an interface to embed Firedrake into ML frameworks.

We build a custom Firedrake operator within \emph{torch.autograd} \citep{paszke_automatic_2017} and JAX \cite{jax2018github} to represent $\mathcal{F}$. The computational graph evaluation and differentiation is delegated to \emph{torch.autograd} and JAX for PyTorch-based and JAX-based nodes, and to \emph{dolfin-adjoint} for Firedrake-based nodes. \newline
% 1 line space
\begin{figure}[ht!]
\centering
\includegraphics[width=.70\linewidth]{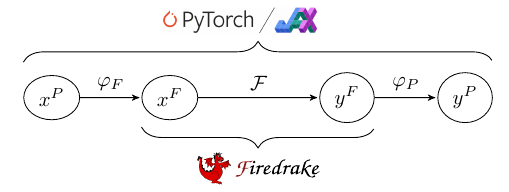}
\caption{Subgraph of the PyTorch/JAX computational graph containing Firedrake operations of interest represented by $\mathcal{F}$, where $P$ refers to PyTorch/JAX variables and $F$ to Firedrake variables. $\varphi_{F}$ and $\varphi_{P}$ represent the casting of a PyTorch/JAX tensor to a Firedrake Function and vice versa.}
\label{fig:coupling_firedrake_pytorch}
\end{figure}

The core data object in each case is merely a tensor of values. In contrast to machine learning frameworks, FEM packages such as Firedrake explicitly associate state tensors with computational domain information. Consequently, the definition of Firedrake tensors induces the required mapping $\varphi_{F}$ and $\varphi_{P}$ to convert to and from PyTorch, see appendix \ref{sec:custom_mappings} for more details. Another consequence is that the Firedrake tensor representation facilitates the use of appropriate inner products, such as $L^p$ and $H^p$,  in addition to the $\ell^{p}$-norm ubiquitous in machine learning. This is critical to guarantee mesh independence for PDE-constrained optimisation problems \citep{schwedes_mesh_2017}.\newline

Listing \ref{code:coupling_template} demonstrates how one can build physics-driven models using PyTorch and Firedrake. Coupling both frameworks only requires a one-line change to existing codes as highlighted in listing \ref{code:coupling_template} (line 9), which makes the implementation straightforward from a user perspective. Line 7 defines the functional $\mathcal{F}$ as a function of given control(s), which enables to only traverse the relevant part of $\mathcal{F}$'s computational graph needed to differentiate $\mathcal{F}$ with respect to the given control(s).  We also extend dolfin-adjoint to relax the assumption that it owns the final quantity of interest that is differentiated, thereby enabling backpropagation computation to start from a PyTorch tensor. The operator $G$ defined in line 9 wraps the operator $F$ to act on PyTorch tensors as illustrated in line 11 with $x^{P}$ and $y^{P}$.

\begin{listing}[ht]
\captionof{listing}{Outline of backpropagation through Firedrake using $\operatorname{fem\_operator}$. The ML backend is chosen using the module import. For JAX, one just needs to import from $\operatorname{firedrake.ml.jax}$ instead.}
\label{code:coupling_template}
\begin{mintedbox}[highlightlines={9}]{python}
import torch
import firedrake as fd
import firedrake.ml.pytorch as fd_ml
import firedrake.adjoint as fda
...
# Defined reduced functional |$\mathcal{F}$| with respect to control(s)
F = fda.|\textcolor{blue}{ReducedFunctional}|(y_F, |\textcolor{blue}{Control}|(x_F))
# Define the coupling operator: |$G \vcentcolon= \varphi_{P} \circ \mathcal{F} \circ \varphi_{F}$|
G = fd_ml.|\textcolor{blue}{fem\_operator}|(F)
# Apply the coupling operator to a torch.Tensor |$x^{P}$|
y_P = G(x_P)
# Backpropagate through |$G$|: calculate |$\mathcal{J}^{*}_{y^{P}, x^{P}}(w^{P})$|
w_P = ...
y_P.backward(w_P)
\end{mintedbox}
\end{listing}
\section{Experiments}
\label{sec:experiments}

We showcase our interface on different examples pertaining to different physics. The ability to embed Firedrake into ML frameworks is showcased in sections \ref{sec:exp:learning_fluid_div_free_reg} and \ref{sec:exp:heat_conductivity}, while the ability to embed ML models in Firedrake is illustrated in section \ref{sec:exp:seismic_inversion}. Finally, section \ref{sec:exp:learn_const_models} demonstrates the ability to combine both types of embeddings.

\subsection{Learning fluid flows with divergence-free regularisation}
\label{sec:exp:learning_fluid_div_free_reg}

In this example, we consider a fluid flow past a fixed circular cylinder in two dimensions. We aim at learning the operator $u(\cdot, t) \mapsto u(\cdot, t + \Delta t)$ for $t\ge0$ and a timestep $\Delta t >0$, where $u$ is the fluid velocity and can be modelled by the following incompressible Navier-Stokes equation:
\begin{equation}
\label{eq:NS_incompressible}
\begin{aligned}
    \frac{\partial u}{\partial t} - \nabla \cdot 2\nu \varepsilon(u) + (u \cdot \nabla) u + \nabla p &= f \quad \text{ in } \Omega \times (0, T] \\
    \nabla \cdot u &= 0 \quad \text{ in } \Omega \times [0, T] \\
    u(\cdot, t=0) &= u_{0} \quad \text{ in } \Omega\\
\end{aligned}    
\end{equation}
where $\varepsilon(u) = \frac{1}{2}(\nabla u + \nabla u^T)$, $\nu > 0$ is the viscosity coefficient, $f$ is the source term, and $u_0$ is the initial condition. We also equip \eqref{eq:NS_incompressible} with the boundary conditions corresponding to this test case as defined in \cite{jackson_finite-element_1987}, with a different inflow condition. More details can be found in appendix \ref{sec:cylinder_flow_appendix}.

In many realistic cases, the Navier-Stokes problem \eqref{eq:NS_incompressible} provides an incomplete description of the physics of interest. Consequently, enforcing the PDE when learning the fluid flow may result in undesired effects. On the other hand, the incompressibility of the fluid, modelled by the divergence-free condition ($\nabla \cdot u = 0$), is a property that should be satisfied by a machine learning method. 

We propose a loss function that uses a weighted $H(div, \Omega)$-norm, where the space $H(div, \Omega)$ consists in vector fields in $L^2(\Omega)$ with divergence in $L^2(\Omega)$:
\begin{equation}
    \mathcal{L} = \frac{1}{N} \sum_{i=1}^{N}\| u_{i} - u^{*}_{i}\|_{H(div, \Omega)}^{2} = \frac{1}{N} \sum_{i=1}^{N}\left(\| u_{i} - u^{*}_{i}\|_{L^2(\Omega)}^{2} + \alpha \| \nabla \cdot u_{i} - \nabla \cdot u^{*}_{i}\|_{L^2(\Omega)}^{2}\right)
\end{equation}
with $\alpha > 0$, $u_{i} = u(\cdot, i\Delta t)$ is the predicted velocity, and $u^{*}_{i} = u^{*}(\cdot, i\Delta t)$ is the velocity obtained by solving \eqref{eq:NS_incompressible} with the finite element method. Given that $u^{*}$ satisfies the divergence-free condition in the weak sense, i.e. $\int_{\Omega} \nabla \cdot u^{*}\,q \diff x = 0$ for any $q \in L^{2}(\Omega)$, we have
\begin{equation}
    \label{eq:NS_div_free_regularisation}
    \mathcal{L} = \frac{1}{N} \sum_{i=1}^{N} \left(\| u_{i} - u^{*}_{i}\|_{L^2(\Omega)}^{2} + \alpha \| \nabla \cdot u_{i}\|_{L^2(\Omega)}^{2}\right)
\end{equation}
where $\alpha > 0$ is the regularisation coefficient. This functional can be understood as an $L^2$-error with a divergence-free regularisation that guides the training to satisfy the incompressiblity of the fluid. Our interface automates the evaluation of this loss, which is achieved by assembling the integrals involved in equation \eqref{eq:NS_div_free_regularisation} using the finite element method over the mesh considered. Listing \ref{code:div_free} in appendix \ref{sec:cylinder_flow_appendix} illustrates the simplicity by which this type of regularisation can be written with our interface.

We consider a graph network architecture, similar to \cite{pfaff_learning_2021}, to learn the one-step simulator. The learned model can be applied iteratively to generate long trajectories at inference time. To the best of our knowledge, this type of complex differentiable physical constraint has not been proposed before and illustrates the range of applications that our interface unlocks.

\begin{figure}[htp]
\centering
\includegraphics[width=.8\linewidth]{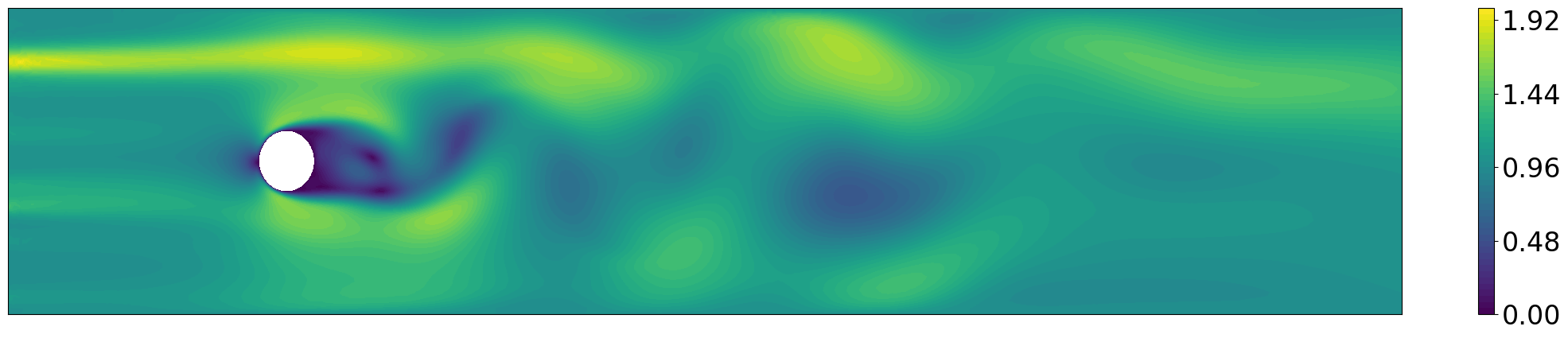}
\includegraphics[width=.8\linewidth]{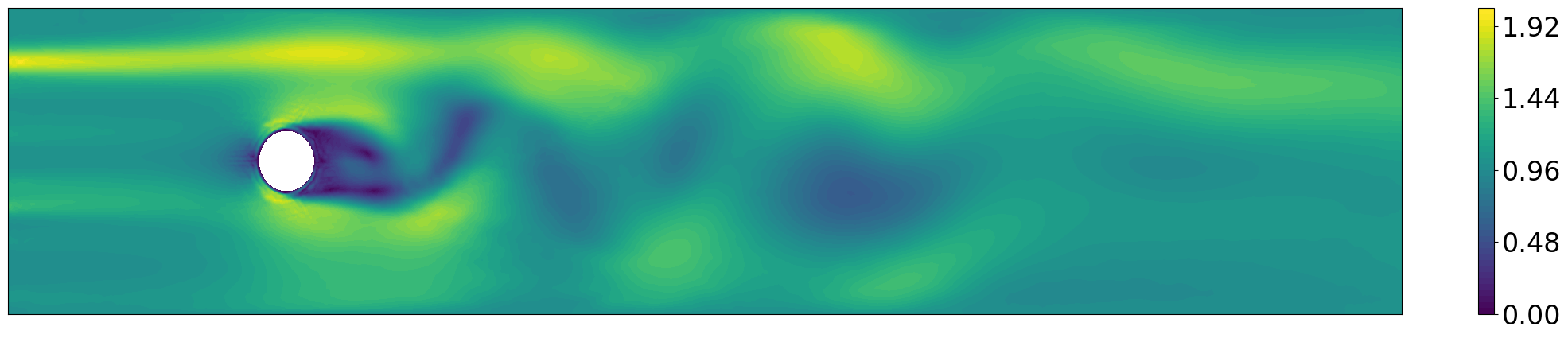}
\caption{Velocity magnitude of $u^{*}$ (top) and $u$ (bottom) at a final time $T = 30$ for $\nu = 5.10^{-3}$.}
\label{fig:cylinder_flow_results}
\end{figure}

\subsection{Differentiable solvers for learning inverse problems}
\label{sec:exp:heat_conductivity}

We consider a simple heat conductivity problem modelled via the following stationary heat equation: Find the temperature field $u \in V$ such that:
\begin{equation}
    \label{eq:heat_conductivity_strong_form}
    \begin{aligned}
        - \nabla \cdot \left( e^{\kappa} \nabla u\right) &= f \quad \text{in } \Omega\\
        u &= 0 \quad \text{on } \partial \Omega
    \end{aligned}
\end{equation}
where $\Omega \subset \reels^{2}$ is an open and bounded domain, $\kappa \in V$ is the conductivity, $f \in V$ the source term, and with $V$ a suitable function space (i.e. $V = H^{1}_{0}(\Omega)$). We want to solve the inverse problem driven by \eqref{eq:heat_conductivity_strong_form} by learning the inverse operator:
\begin{equation}
    \label{eq:inverse_operator}
    \hat{\kappa}: u^{obs} \rightarrow \kappa
\end{equation}
where $u^{obs}$ refers to an observed temperature field and is modelled by: $u^{obs} = u(\kappa) + \varepsilon$, with $u(\kappa)$ the solution of \eqref{eq:heat_conductivity_strong_form} for a given $\kappa$, and $\varepsilon$ the observation noise. We use a model-constrained deep neural network approach as in \cite{nguyen_tnet_2022}, where we embed a PDE solver for \eqref{eq:heat_conductivity_strong_form} in the loss. More precisely, let $\kappa_{\theta}$ be a a neural network based model with $\theta$ the model parameters. We train $\kappa_{\theta}$ to learn the inverse operator defined in \eqref{eq:inverse_operator} by considering the following loss function for each batch element:
\begin{equation}
    \label{eq:tnet_J}
    \mathcal{L} = \frac{1}{2}\|\kappa_{\theta}(u^{obs}) - \kappa^{exact}\|^{2}_{\normltwo(\Omega)} + \frac{\alpha}{2} \|u(\kappa_{\theta}(u^{obs})) - u^{obs}\|_{\normltwo(\Omega)}^{2}
\end{equation}
where $\kappa^{exact}$ refers to the exact conductivity for a given observable $u^{obs}$. \cite{nguyen_tnet_2022} employed a single hidden layer model, here we consider a convolutional neural network architecture (CNN) for learning the conductivity. Two Firedrake coupling operators are used during training: one for computing the PDE solution $u(\cdot)$, and the other for assembling the $L^2$-loss as the choice of norm is critical for mesh-independence. For training, we generate $n$ random fields to form the training split: $\{\kappa^{exact}_{i}, u^{obs}_{i}\}_{1 \le i \le n}$, and average the loss defined in \eqref{eq:tnet_J} across training samples. For evaluation, we generate data in the same way and average the error across test samples. More details about this test case can be found in appendix \ref{sec:heat_conductivity_appendix}. 
\begin{figure}[htp]
\centering
\includegraphics[width=.35\linewidth]{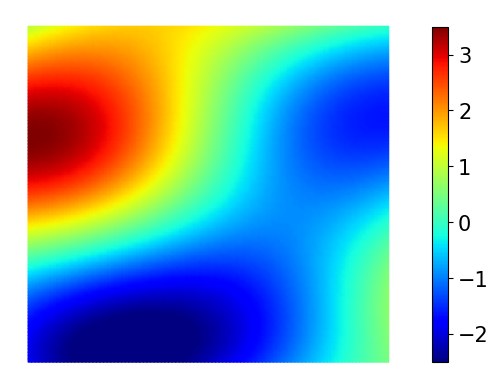}
\hspace{0.5cm}
\includegraphics[width=.35\linewidth]{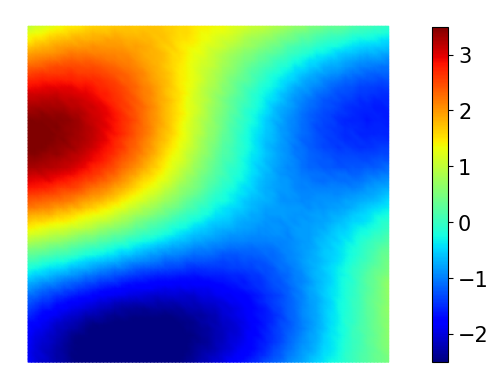}
\caption{Heat conductivity as a function of position $(x, y)$: exact conductivity (left), conductivity reconstructed from observed data $\kappa_{\theta}(u^{obs})$ by the CNN model (right).}
\label{fig:heat_conductivity_models}
\end{figure}

\subsection{Learning constitutive models from experimental data}
\label{sec:exp:learn_const_models}

Constitutive models describe the relationship between deformations and internal forces of materials \citep{Love1920, Hill1950}. They are ubiquitous in material design \citep{Oyen2013, Uszball2023, Bouvard2011, Fang2004, Wang2019, Suo2023}, structural analysis \citep{Chen1995, Cervenka2024}, and safety assessment \citep{Zheng2022, Matos2019}, to name but three. They are functions that model the material behaviour by mapping the strain tensor $\varepsilon(u)$, which quantifies the deformation of the material as a function of displacement u, to the stress tensor $\sigma$, which describes the internal forces within the material. This mapping takes various forms depending on the material considered and the behaviour being modelled (e.g. hyperelasticity). Existing approaches traditionally derive such models via simplified assumptions that result in parametrisations that can be fit to experimental data. Machine learning offers a sophisticated data-driven alternative.

The naive approach consisting in learning these models directly is not viable as the stress tensor is not directly measurable \cite{hamel2022}, unlike experimental data such as the displacement. Instead, we propose to learn constitutive models using signals from the measured displacement. The displacement $u$ is the solution of a PDE, which directly depends on the stress tensor resulting from the constitutive model. This requires the ability to embed the ML model inside the PDE, to calculate the displacement using the predicted stress tensor, but also the ability to embed the PDE solver inside the loss function during training to compute the error with respect to the measured displacement.

We consider the three-point bending test \citep{ASTMD790_17}, a standard method for characterising the mechanical properties of materials. A beam is placed on two supports and is subjected to a force applied at the centre of the top surface, which causes a deflection in the beam. The displacement is modelled by the following PDE:
\begin{equation}
    \label{eq:PDE_const_model}
    \begin{aligned}
    \nabla \cdot \sigma_{\theta}(\varepsilon(u)) &= 0  \quad \text{in } \Omega \\
    u_y &= 0 \quad \text{on } \partial\Omega_S \\
    \sigma_{\theta}(\varepsilon(u)) \cdot n &= f  \quad \text{on } \partial\Omega_P
    \end{aligned}
\end{equation}
where $\sigma_{\theta}$ is the constitutive model being learned, $u_y$ is the vertical component of the displacement $u$, $n$ is the outward-facing unit normal, and $f$ is the traction corresponding to the applied load. $\partial \Omega_{P}$ refers to the surface where the force is applied and $\partial \Omega_{S}$ to the surface of the beam in contact with the two supports.
A synthetic experimental force-deflection dataset is used to train the ML constitutive model. More details can be found in appendix \ref{sec:ml-based_constitutive_details}.
\begin{figure}[h!]
    \centering
    \begin{subfigure}[b]{0.45\textwidth}
        \centering
        \includegraphics[width=\textwidth]{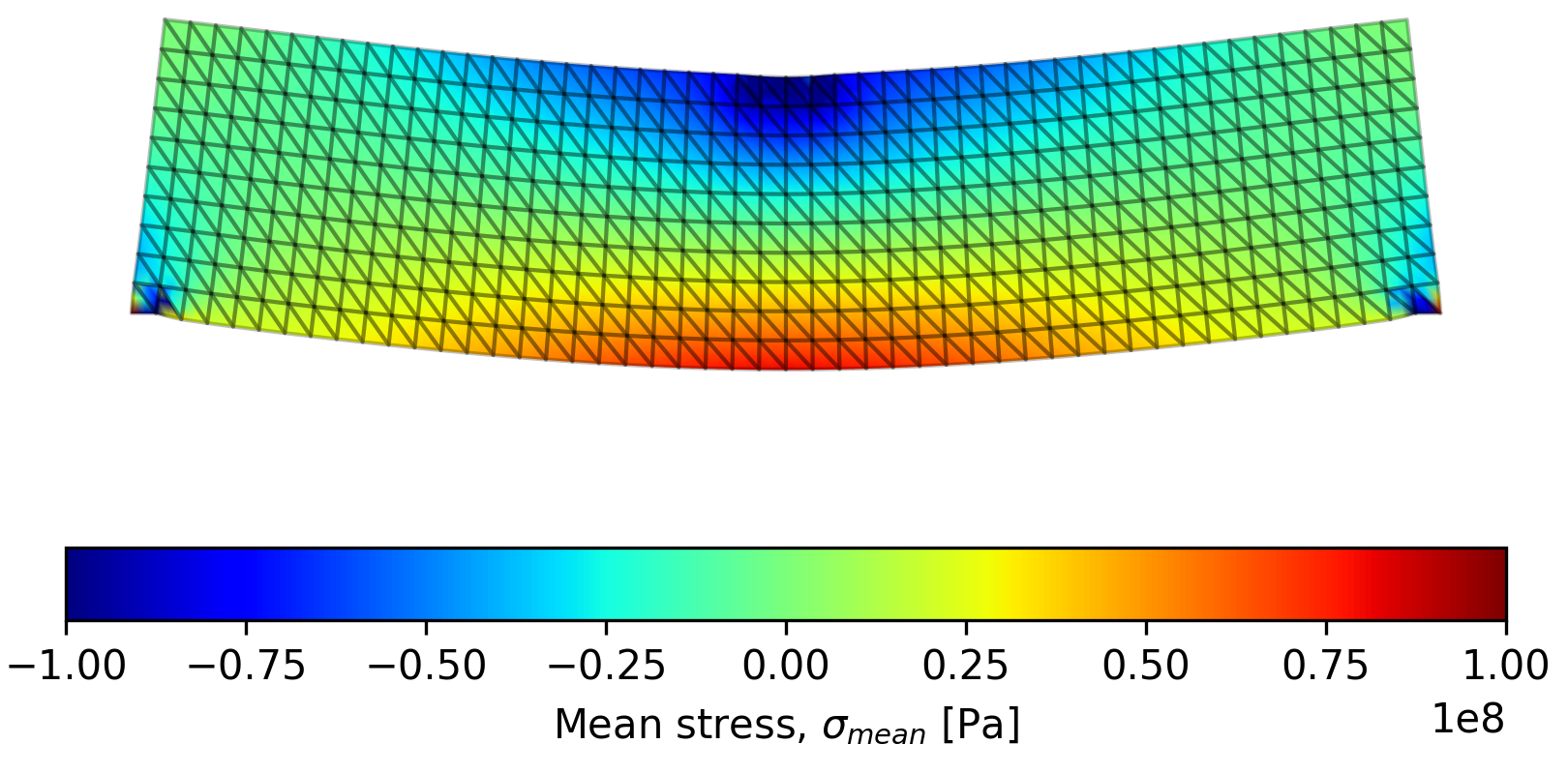}
        \caption{}
        \label{fig:mean_stress_exact}
    \end{subfigure}
    \hfill
    \begin{subfigure}[b]{0.45\textwidth}
        \centering
        \includegraphics[width=\textwidth]{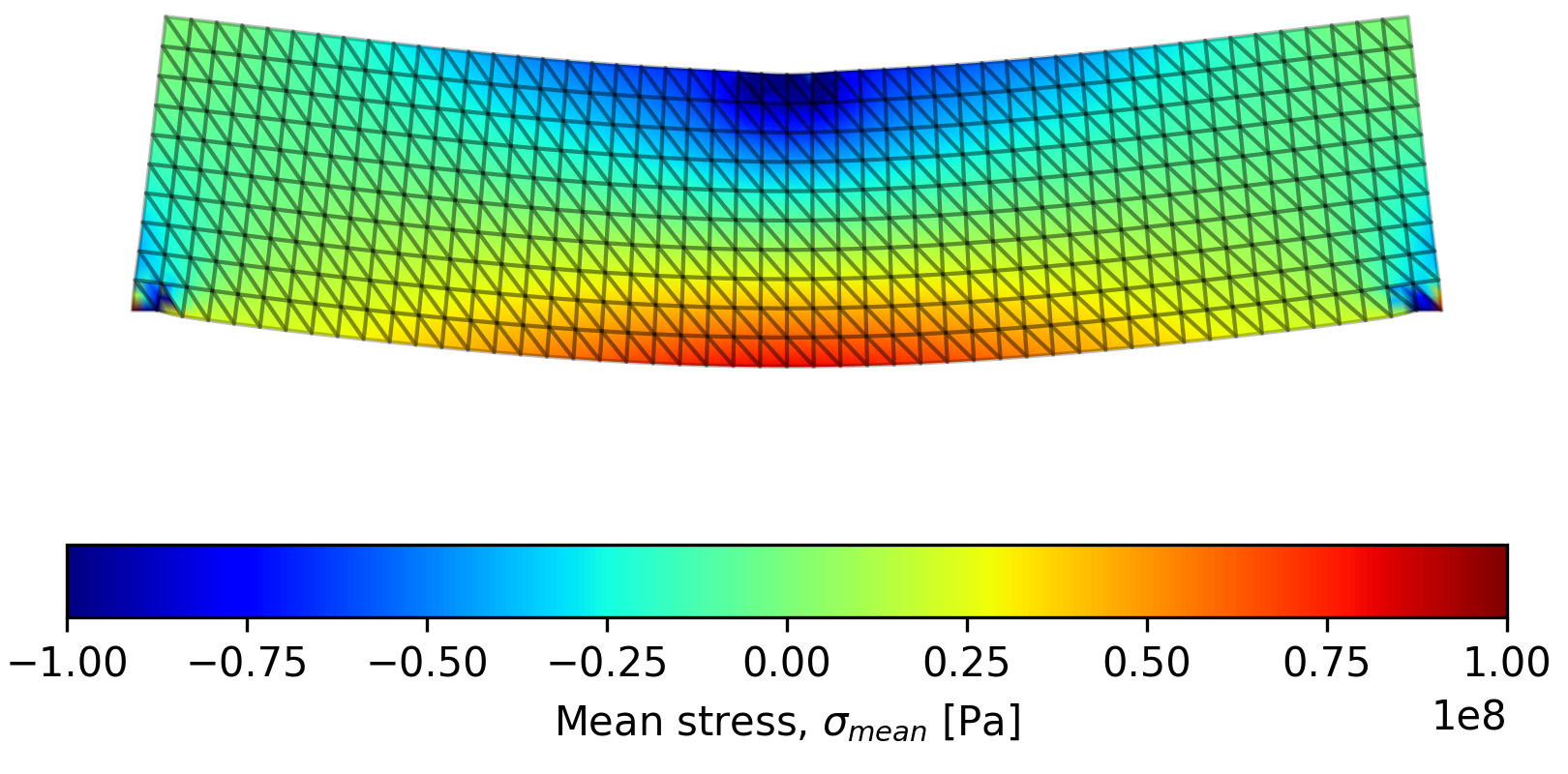}
        \caption{}
        \label{fig:mean_stress_ml}
    \end{subfigure}
    \caption{Mean stress tensor of the exact constitutive model (a) and the ML constitutive model (b)}
    \label{fig:tpb_results}
\end{figure}

To the best of our knowledge, the generic training of constitutive models embedded in finite-element PDE models, which involves differentiation with respect to both the model parameters and the PDE solution, has not been proposed before.

\subsection{ML-based regularisers for seismic inversion}
\label{sec:exp:seismic_inversion}

Seismic inversion is the use of seismic reflection data to infer the material properties of the Earth's subsurface. The problem arises predominantly in exploration geophysics (e.g. oil and gas prospection) \cite{Multiscale_FWI_2013, Seismic_Imaging_adjoing_tomography_2012, Seismic_Tomography_2010} and geotechnical site characterisation \cite{FWI_site_characterization_2013}. This problem can be formulated as an inverse wave problem and is invariably ill-posed: various solution artifacts result in indistinguishable functionals but nonphysical solutions. A simple model problem can be formulated as follows:
\begin{equation}
\label{regularizer_example}
\min_{c \in P}\ \ \frac{1}{2}\|\varphi(c) - \varphi^{obs}\|_{V}^{2} + \alpha \mathcal{R}(c)
\end{equation}
where $\varphi^{obs}$ are the observed data, $c$ the scalar wave speed, $\alpha$ the regularisation factor and $\varphi \in V$ is the wave displacement of the medium such that:
\begin{equation}
\label{regularizer_wave_Example}
\begin{aligned}
  \frac{\partial^{2} \varphi}{\partial t^{2}}  - c^{2} \Delta \varphi &= f \quad \textrm{in}\ \Omega\\
 	\varphi(t=0) &= \varphi_0.
\end{aligned}
\end{equation}
Equation \eqref{regularizer_wave_Example} is referred to as the \emph{forward problem}. In practice, one may use a more complex formulation of this equation to take into account additional physical effects such as elasiticy, acousticity or anisotropy. More realistic applications may also have complex boundary conditions.

Equation \eqref{regularizer_example} is the function to optimise. The first term accounts for the mismatch error between the solution obtained by the forward problem (the PDE) and the observed data (the \emph{fidelity term}). $\mathcal{R}$ is a regularisation operator which attempts to counter noisy observations and to make the problem well-posed. The PDE is based on fundamental physical laws but the regulariser is more heuristic in nature. A number of recent works have employed deep learning to build regularisers for inverse problems, especially for seismic inversion \cite{zhang_chapter_2022, shi_deep_2020} and medical imaging \cite{lunz_adversarial_2018, li_nett_2019, adler_solving_2017}. ML operators in UFL provide a high-level programming abstraction for this problem. The cost function simply contains an ML operator representing the neural network model, i.e. $\mathcal{R}(c) = \frac{1}{2}\|N(c)\|^{2}$, with $N$ an ML operator. Listing \ref{code:seismic-inversion} in appendix \ref{sec:seismic_inversion_appendix} illustrates this in Firedrake.

\begin{figure}[htp]
\centering
\includegraphics[width=.27\linewidth]{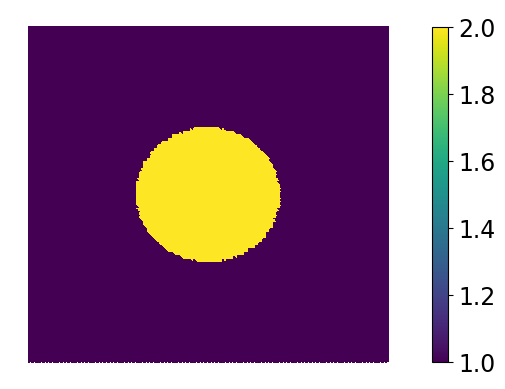}
\includegraphics[width=.29\linewidth]{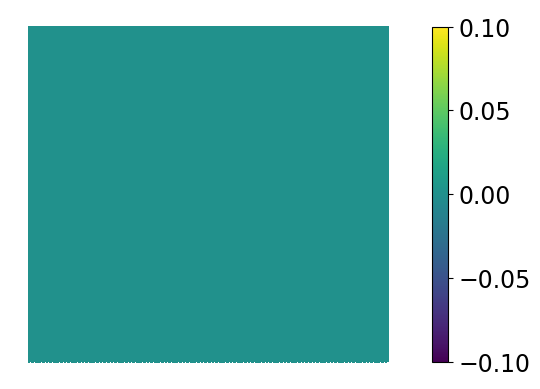}
\includegraphics[width=.27\linewidth]{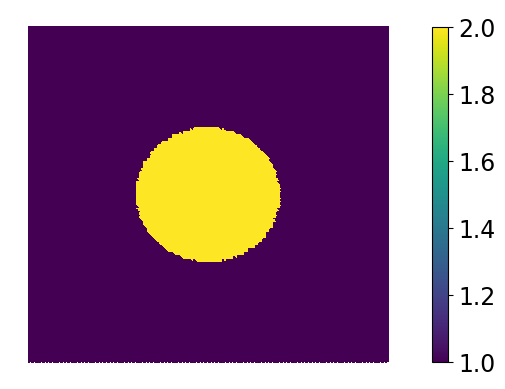}
\caption{Recovered wave speed $c$ as a function of position ($x,z$) obtained for the model \eqref{regularizer_example}-\eqref{regularizer_wave_Example}: exact velocity (left), Tikhonov regulariser (middle), neural network-based regulariser (right).}
\label{fig:velocity_models_seismic_inversion}

\end{figure}

\section{Conclusion}
\label{sec:conclusion}

In this paper, we introduce a differentiable programming interface that couples PDE-based systems implemented in Firedrake with machine learning models implemented in PyTorch and JAX to yield end-to-end differentiable systems with state-of-the-art capabilities. Our interface extends the range of applications currently possible in the literature and unlocks new research directions.

\textbf{Limitations} Our framework involves finite-element modelling of the PDE part. Consequently, collocation approaches such a physics-informed neural networks (PINNs) cannot directly be employed in our setting and other dedicated software can be used \cite{lu_deepxde_2019, coscia_pina_2024}. However, our framework can be used for PINNs with a finite-element residual in the loss.

\subsection*{Acknowledgments}
  Nacime Bouziani was supported by the Eric and Wendy Schmidt AI in Science Postdoctoral Fellowship, a Schmidt Futures program.

\bibliographystyle{plain}
\bibliography{main}

%%%%%%%%%%%%%%%%%%%%%%%%%%%%%%%%%%%%%%%%%%%%%%%%%%%%%%%%%%%%

\newpage
\appendix

\section{Appendix / supplemental material}

\subsection{Adjoint model of PDE solutions}
\label{sec:adjoint_method}

Let $V$ and $M$ be Hilbert spaces, and let $u \in V$ be the solution of the parametrised partial differential equation defined as:
\begin{equation}
    \label{eq:appendix:adjoint:F_residual}
    F(u, m;  v) = 0 \quad \forall v \in V
\end{equation}

with $F$ the variational form of the PDE, and $m \in M$ the control value. We assume that the PDE defined by $F$ yields a unique solution for any control value $m \in M$. Hence, we can define the PDE solution operator $u(\cdot) \vcentcolon M \rightarrow V$ that maps any control value $m$ to the corresponding solution $u(m)$ satisfying \eqref{eq:appendix:adjoint:F_residual}. Note that the PDE solution $u(m)$ is generally not known explicitly but can be approximated using the finite element method. \\

We assume that the linear form $F$ is continuously Fr\'echet differentiable and that the linearised PDE operator $\frac{\partial F}{\partial u}$ is invertible. It follows, using implicit function theorem, that the solution operator $u(\cdot)$ is continuously Fr\'echet differentiable, see \citep[Sec. 1.4.2]{hinze_optimization_2009}. Let $V^{*}$ and $M^{*}$ be the dual spaces of $V$ and $M$, i.e. the spaces of all continuous linear functionals on $V$ and $M$, respectively. Backpropagating through the PDE solution $u(m)$ equates to computing its adjoint model $\mathcal{J}^{*}_{u, m} \vcentcolon V^{*} \rightarrow M^{*}$ defined as:
\begin{equation}
    \label{eq:appendix:adjoint:adj_model_def}
    \mathcal{J}^{*}_{u, m}(w) = \frac{du}{dm}^{*} w \quad \forall w \in V^{*}
\end{equation}

where $\frac{du}{dm}^{*} \in \mathcal{L}(V^{*}, M^{*})$ is the adjoint of the G\^ateaux derivative of $u$ with respect to $m$. While the expression of $\frac{du}{dm}^{*}$ is not known explicitly, given that $u(m)$ is the PDE solution, we can still compute it using \eqref{eq:appendix:adjoint:F_residual}. More precisely, given that we have $F(u(m), m; v) = 0$ for all $m \in M$, it follows that the derivative $\frac{dF}{dm}$ must be zero. Therefore, applying chain rule yields:
\begin{equation}
    \label{eq:appendix:adjoint:dFdm_0}
    \frac{\partial F}{\partial u} \frac{du}{dm} + \frac{\partial F}{\partial m} = 0
\end{equation}

which leads to
\begin{equation}
    \label{eq:appendix:adjoint:dudm}
    \frac{du}{dm} = - \frac{\partial F}{\partial u}^{-1} \frac{\partial F}{\partial m}
\end{equation}

since we assumed $\frac{\partial F}{\partial u}$ to be invertible. Combining equations \ref{eq:appendix:adjoint:dudm} and \ref{eq:appendix:adjoint:adj_model_def}, we finally obtain for all $w \in V^{*}$:
\begin{equation}
    \label{eq:appendix:adjoint:adj_model}
    \mathcal{J}^{*}_{u, m}(w) = - \frac{\partial F}{\partial m}^{*} \lambda
\end{equation}

where $\lambda \in V$ is the solution of the adjoint equation defined as
\begin{equation}
    \label{eq:appendix:adjoint:adj_equation}
    \frac{\partial F}{\partial u}^{*} \lambda = w
\end{equation}
%Therefore, when backpropagating tAdd link backprop call example u(kappa theta) ... 

We refer the interested reader to \citep{schwedes_mesh_2017}, \citep[Sec. 1.6]{hinze_optimization_2009}, and \citep{bouziani_differentiable_2023} for more details. %on the adjoint of PDE-constrained problems.

\begin{example}
Let $V$ and $M$ be a suitable function spaces and $f$ a given function in $M$. The modified Helmholtz problem is given by: find $u \in V$ such that: 

\begin{equation}
\label{example:var_form}
\int_{\Omega} u\, v + \nabla u \cdot \nabla v\ \diff x - \int_{\Omega} f\, v \diff x = 0 \quad \forall v \in V.
\end{equation}

The adjoint model of the solution $u$ of this simple PDE with respect to the right-hand side $f$ can be computed in a few lines of code in Firedrake as illustrated in listing \ref{code:Assemble-Poisson}. We consider $\Omega \vcentcolon = [0, 1]^{2}$ and a corresponding mesh with 50 cells in each direction, see line 7. In lines 8 and 9, we discretise the spaces $M$ and $U$ using Lagrange polynomials of degree $1$ and $2$, respectively. The linear PDE is solved in line 21 using $LU$ factorisation. Finally, the adjoint model $\mathcal{J}^{*}_{u, f}$ is automatically calculated in line 25. This single line of code will solve the adjoint equation \eqref{eq:appendix:adjoint:adj_equation}, and compute the adjoint model using \eqref{eq:appendix:adjoint:adj_model}.
\newpage
\begin{listing}
\captionof{listing}{Firedrake code solving the variational problem \eqref{example:var_form} and computing the adjoint model $\mathcal{J}^{*}_{u, f}$. Note the similarity between line 18 and \eqref{example:var_form}.}
\label{code:Assemble-Poisson}
\begin{mintedbox}[highlightlines={25}]{python}
from firedrake import *
from firedrake.adjoint import *

# Start annotation
continue_annotation()

mesh = |\textcolor{blue}{UnitSquareMesh}|(50, 50)
M = |\textcolor{blue}{FunctionSpace}|(mesh, "Lagrange", 1)
V = |\textcolor{blue}{FunctionSpace}|(mesh, "Lagrange", 2)
u = |\textcolor{blue}{Function}|(V)
v = |\textcolor{blue}{TestFunction}|(V)

# Define the right-hand side
x, y = |\textcolor{blue}{SpatialCoordinate}|(mesh)
f = |\textcolor{blue}{Function}|(M).interpolate(2 * pi**2 * sin(pi * x) * sin(pi * y))

# Residual form
F = (u * v + |\textcolor{blue}{inner}|(|\textcolor{blue}{grad}|(u), |\textcolor{blue}{grad}|(v))) * dx - f * v * dx

# Solve the PDE
|\textcolor{blue}{solve}|(F == 0, u, solver_parameters={"ksp_type": "preonly", "pc_type": "lu"})

# Compute |$\mathcal{J}^{*}_{u, f}(w)$|
w = ...
u_adj = |\textcolor{blue}{compute\_gradient}|(u, |\textcolor{blue}{Control}|(f), adj_value=w)
\end{mintedbox}
\end{listing}
\end{example}

\subsection{Machine learning operators in UFL and Firedrake}
\label{sec:ml_operator_firedrake_ufl}

The symbolic representation of finite element problems in UFL \cite{UFL_2014} is organised around representing parametrised multilinear forms, and in particular variational forms. A parametrised \emph{multi-linear form} (or \emph{linear n-forms}) is a map from the product of a given sequence $\left\lbrace W_{j} \right\rbrace_{j=1}^{k} \times \left\lbrace V_{j} \right\rbrace_{j=0}^{n-1}$ of function spaces to a field $K$ ($\reels$ or $\complex$):
\begin{equation}
\label{def:multilinear_form}
W_{1} \times W_{2} \times \cdots \times W_{k} \times V_{n-1} \times V_{n-2} \times \cdots \times V_{0} \rightarrow K.
\end{equation}
that is linear in each argument on the \textit{argument spaces} $\left\lbrace V_{j} \right\rbrace_{j=0}^{n-1}$. The spaces $\left\lbrace W_{j} \right\rbrace_{j=1}^{k}$ are traditionally labelled as \textit{coefficient spaces}. The number of argument spaces is referred to as the \emph{arity} of a linear $n$-form. Variational forms with arity $n = 0,\ 1,\ $ and $2$ are respectively named \emph{functionals}, \emph{linear forms} and \emph{bilinear forms}. These can be assembled to produce a scalar, a vector and a matrix, respectively.

We refer to \eqref{def:multilinear_form} as a linear $n$-form with $k$ coefficients. While the multilinear form is necessarily linear in each argument, it can be nonlinear in each coefficient function. An argument is an unknown function in a finite element space, while a coefficient is a known function in a finite element space.  UFL has a symbolic abstraction to represent variational forms as linear $n$-forms with $k$ coefficients, that is defined as a sum of integrals over subspaces of the problem domain.

The representation of a machine learning operator naturally fits within the UFL representation of a linear form, as any operator between Hilbert spaces induces a corresponding linear form. More precisely, a machine learning operator is represented as a linear form with $k$ coefficients corresponding to the inputs:
\begin{equation}
\label{def_external_operator_FORM}
\begin{aligned}
N \vcentcolon  W_1 \times \cdots \times W_k \times \mathcal{X}^{*} &\longmapsto K \\
 u_1,\ \ldots,\ u_k,\ v^{*} \quad &\longrightarrow N(u_1, \ldots, u_k; v^{*})
\end{aligned}
\end{equation}
with $X^{*}$ the dual space of $X$, and where the linear form $N(u_1, \ldots, u_k; v^{*})$ is linear with respect to $v^{*}$ but can be nonlinear with respect to its operands $(u_{i}){1\le i\le k}$. Expressing machine learning operators as linear forms facilitates their symbolic and numerical composition with the variational forms of PDEs.

Our interface comprises a symbolic representation of machine learning operators, as well as a numerical implementation that calls the corresponding machine learning framework depending on the required evaluation. The application of symbolic operations such as $\operatorname{action}$ or $\operatorname{adjoint}$ on machine learning operators, as illustrated in listing \ref{code:neural_net_interface}, translates at assembly time into appropriate calls to the ML framework considered.

\subsection{Custom mappings}
\label{sec:custom_mappings}

Depending on the user case, different representations can be adopted to represent finite element functions in the ML framework considered. For example, one can feed a neural network with the values of a finite element function on a set of points, e.g on a uniform Cartesian grid for CNN-based architectures or a more general grid for graph neural networks. This representation is independent of the mappings $\varphi_{F}$ and $\varphi_{P}$ illustrated in figure~\ref{fig:coupling_firedrake_pytorch} which merely cast tensors from one framework to the other without altering their shape. This consideration is rather an operation that happens either in Firedrake and/or PyTorch/JAX. \newline

In particular, complex representations of finite element functions can be lifted at the level of the definition of the mesh and function spaces in Firedrake, as it inherently provides a richer space representation than PyTorch/JAX.

\subsection{Experiment details}
\label{sec:exp_details_appendix}

\subsubsection{Learning fluid flows with divergence-free regularisation}
\label{sec:cylinder_flow_appendix}

We generate the dataset by solving the Navier-Stokes problem \eqref{eq:NS_incompressible} for a final time $T = 30$. We use a Taylor-Hood discretisation in space and a Crank-Nicholson scheme for time discretisation. We record the velocity solutions every $\Delta t_{s} = 0.2$ to construct our dataset. The training, validation, and test splits are formed by considering solutions $u(\cdot, t)$ for $t \in (0, 20]$, $t \in (20, 25]$, $t \in (25, 30]$, respectively.

We employ a message passing neural network architecture with an encode-process-decode architecture similar to \cite{pfaff_learning_2021}. We learn the one-step simulator $u(\cdot, t) \mapsto u(\cdot, t + \Delta t)$ for a prediction timestep $\Delta t = 2$. We use an exponential learning rate decay from $10^{-4}$ to $10^{-6}$, and perform training for $15000$ epochs using the Adam optimiser. We obtain a relative $L^{2}$ error of $5.6\, 10^{-2}$ on the training split and $1.6\, 10^{-1}$ on the test split. This can be explained by the different flow patterns in both splits, as the vortex structures start to form at the end of the training split and are prevalent in the test split, i.e. for $t > 25$. Other methods can be employed to stabilise the roll-out and yield better performance \cite{brandstetter_message_2023, lippe_pde-refiner_2023}.

\begin{figure}[h!]
    \centering
    \includegraphics[width=1.\textwidth]{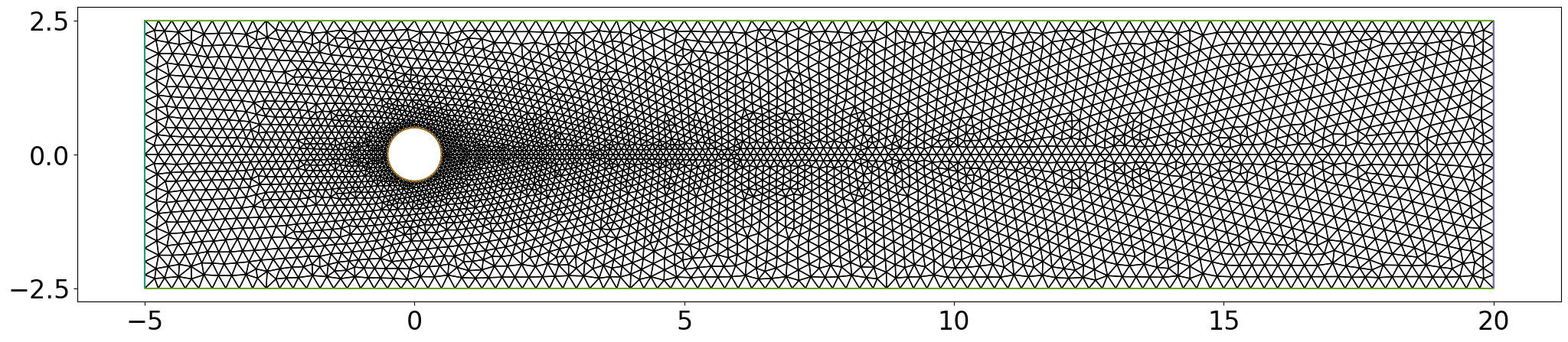}
    \caption{Mesh considered for the Navier-Stokes problem \eqref{eq:NS_incompressible} and composed of 9068 triangles.}
    \label{fig:mesh_NS_cylinder_flow}
\end{figure}

We consider a rectangular domain $\Omega \vcentcolon= [-5, 20] \times [-2.5, 2.5]$ with a circle of radius $0.5$ and centered in $(0, 0)$ to represent the cylinder obstacle. The corresponding mesh is generated using Gmsh \cite{geuzaine_gmsh_2009} and illustrated in figure \ref{fig:mesh_NS_cylinder_flow}.

Listing \ref{code:div_free} illustrates how the divergence-free regularisation can be implemented with our interface. From line 1 to 5, we use Firedrake to evaluate the training loss. We then embed the corresponding operation in PyTorch, see line 9. At training time, backpropagating the loss will require Firedrake to backpropagate through the assembly of the loss, which depends on the output $\operatorname{u\_P}$ of the model. This explains why we parametrise the Firedrake operation by the function $u$ in line 8.
\newpage
\begin{listing}
\captionof{listing}{Outline of the code used to compute the loss with divergence-free regularisation. The loss is computed using the finite element assembly in Firedrake and embedded in PyTorch using $\operatorname{fem\_operator}$. $\operatorname{u\_P}$ refers to the prediction of our graph network model and $\operatorname{ustar\_P}$ the target value, which satisfies the divergence-free condition weakly, as detailed in section \ref{sec:exp:learning_fluid_div_free_reg}.}
\label{code:div_free}
\begin{mintedbox}[highlightlines={9}]{python}
def div_free_reg_loss(u, ustar):
    |$\alpha$| = ...
    # Compute |$\norm{u_{i} - u^{*}_{i}}_{L^2(\Omega)}^{2} + \alpha \norm{\nabla \cdot u_{i}}_{L^2(\Omega)}^{2}$|
    L = (|\textcolor{blue}{inner}|(u - ustar, u - ustar) + |$\alpha$| * |\textcolor{blue}{inner}|(|\textcolor{blue}{div}|(u), |\textcolor{blue}{div}|(u))) * dx
    return |\textcolor{blue}{assemble}|(L)

with set_working_tape():
    F = |\textcolor{blue}{ReducedFunctional}|(div_free_reg_loss(u, ustar), |\textcolor{blue}{Control}|(u))
    G = |\textcolor{blue}{fem\_operator}|(F)

...
# -- Training loop -- #
...
# Compute the loss for a batch element
loss = G(u_P, ustar_P)
...
\end{mintedbox}
\end{listing}

\subsubsection{Differentiable solvers for learning inverse problems}
\label{sec:heat_conductivity_appendix}

We generate a synthetic dataset $\{\kappa^{exact}_{i}, u^{obs}_{i}\}_{1 \le i \le n}$ for training and test using the following procedure:

\begin{itemize}
    \item Randomly generate parameter of interest $\{\kappa_{i}\}_{1 \le i \le n}$
    \item Compute for each $\kappa_{i}$ the corresponding solution of the forward problem (PDE): $u(\kappa_{i})$
    \item Add noise to form the observables: $u_{i}^{obs} = u(\kappa_{i}) + \varepsilon_{i} \quad\forall i \in [|1, n|]$, e.g. $\varepsilon_{i} \in \mathcal{N}(0, 1)$.
\end{itemize}

This procedure is not specific to a given PDE and can be adapted to a wide range of inverse problems. For the heat conductivity example, we generate 500 training samples and 100 test samples. For training, we average the loss defined in \eqref{eq:tnet_J} over the $n$ training samples, i.e. we have:

\begin{equation}
    \label{eq:total_loss}
    \mathcal{L} = \frac{1}{2n} \sum_{i=1}^{n} \left( \|\kappa_{\theta}(u^{obs}_{i}) - \kappa^{exact}_{i}\|^{2}_{\normltwo(\Omega)} + \alpha\|u(\kappa_{\theta}(u^{obs}_{i})) - u^{obs}_{i}\|_{\normltwo(\Omega)}^{2} \right)
\end{equation}

For evaluation, we adopt the evaluation metric used in  \cite{nguyen_tnet_2022}, i.e. we report the average relative error on the $m$ test samples:

\begin{equation}
    \label{eq:evaluation_metric}
    R = \frac{1}{m} \sum_{i=1}^{m} \frac{\|\kappa_{\theta}(u^{obs}_{i}) - \kappa^{exact}_{i}\|^{2}_{L^{2}(\Omega)}}{\|\kappa^{exact}_{i}\|^{2}_{L^{2}(\Omega)}}
\end{equation}

The heat conductivity example is mostly designed for demonstrating the criticality of coupling PyTorch/JAX and Firedrake to design, implement, and run complex physics-driven machine learning models in a highly productive way. Despite neither having a baseline to compare with nor an available dataset to report the performance of our trained CNN architecture, we release the dataset we generated for this example for sake of future comparisons. On this synthetic dataset, the average relative $L^2$-error $R$ on the test split is $17.68\%$. This result is not directly comparable with the analogous example presented in \cite{nguyen_tnet_2022} as the dataset and the problem formulation differ. However, given the close similarity between both test cases and the significant gap in performance in favour of our model, we argue that this can be explained by the fact that we have considered a more complex architecture. \newline

\subsubsection{Learning constitutive models from experimental data}
\label{sec:ml-based_constitutive_details}

For this experiment, the linear elasticity setting is considered to generate our dataset, as in this setting the stress is known analytically, which allows direct comparison of the prediction of our model with the exact stress tensor.

The three-point bending test setup used in this example is taken from the experimental data in \cite{farsi2017}. The tested ceramic beam is $38.09 \ \text{mm}$ in length, $4.64 \ \text{mm}$ in height, and $4.64 \ \text{mm}$ in width. The beam is placed on two supports with a $20 \ \text{mm}$ span. The material properties of the ceramic beam are as follows: Young's modulus is 142.29 $GPa$, Poisson's ratio is 0.27, and the density is 3 $g/cm^3$. The synthetic data used for training the ML model is generated by solving the three-point bending test problem with the linear elastic constitutive model. We consider a simple MLP for the ML constitutive model is a simple MLP.

\begin{figure}
    \centering
    \includegraphics[width=0.7\textwidth]{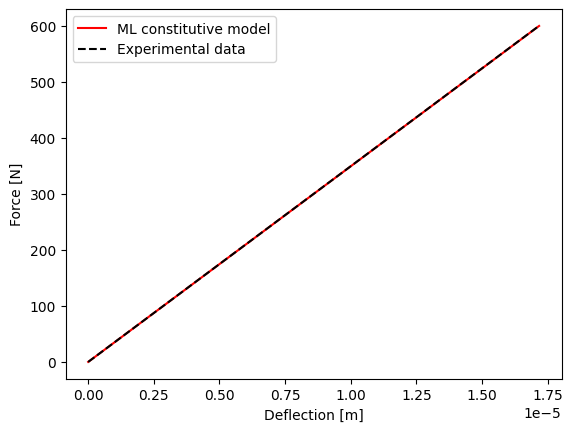}
    \caption{Force-deflection curve corresponding to the synthetically generated experimental data (black dashed) and to the solution of the PDE comprising the embedded ML constitutive model (see \eqref{eq:PDE_const_model})}
    \label{fig:tpb_force-deflection}
\end{figure}

In figure~\ref{fig:tpb_force-deflection}, the black dashed line is the synthetic force-deflection curve of a three-point bending test, which is used to form the splits of our dataset. The red line shows the force-deflection curve of the beam computed with the PDE system containing the trained ML constitutive model. We train for $100$ epochs using the Adam optimiser.

For the training objective, we consider the relative error between the beam deflection $\delta(f_i)$, which corresponds to the evaluation of the displacement at a specific point, and the experimental beam deflection $\delta^{\text{exp}}(f_i)$. In line with the literature, we consider $\delta(f_i) = u(x_j, y_j)$, where $x_j$ and $y_j$ correspond to the middle of the bar. Note that $u$ is the solution of the PDE, and depends on the ML constitutive model, as detailed in \eqref{eq:PDE_const_model}.
\begin{equation}
    \mathcal{L} = \frac{1}{N} \sum_{i=1}^{N} \frac{\mid \delta(f_i) - \delta^{\text{exp}}(f_i) \mid}{\mid \delta^{\text{exp}}(f_i)\mid}
    \label{eq:loss_function}
\end{equation}
The traction $f_{i}$ can be written as $f=\frac{F_{i}}{A_{P}}$, where $F_{i}$ is the force applied and $A_{P}$ the surface where the traction is applied on $\partial \Omega_{P}$.

\subsubsection{ML-based regularisers for seismic inversion}
\label{sec:seismic_inversion_appendix}

For this experiment, we solve the PDE-constrained optimisation problem \eqref{regularizer_example}-\eqref{regularizer_wave_Example} using a pretrained model as a regulariser. We trained a simple MLP model in a supervised manner for a set of different velocities. Listing \ref{code:seismic-inversion} outlines how this problem was solved using our interface in Firedrake.

\newpage
\begin{listing}[h!]
\captionof{listing}{Outline of seismic inversion using a neural network regulariser implemented as an ML operator in Firedrake.}
\label{code:seismic-inversion}
\begin{mintedbox}{python}
from firedrake import *
from firedrake.adjoint import *

...
# Get a pre-trained PyTorch/JAX model
model = ...
# Define the ML operator from the model
ml_op = |\textcolor{blue}{ml\_operator}|(model, function_space=...)
N = ml_op(c)

# Solve the forward problem defined by equation |\eqref{regularizer_wave_Example}|
phi = F(c)
# Assemble the cost function: |$J = \frac{1}{2} \norm{\varphi(c) - \varphi^{obs}}_{L^2(\Omega)}^{2} + \frac{\alpha}{2} \norm{N(c)}_{L^2(\Omega)}^{2} $|
J = |\textcolor{blue}{assemble}|(0.5*(|\textcolor{blue}{inner}|(phi-phi_obs, phi-phi_obs) +
     					   alpha*|\textcolor{blue}{inner}|(N, N))*dx)

# Optimise the problem
Jhat = |\textcolor{blue}{ReducedFunctional}|(J, |\textcolor{blue}{Control}|(c))
c_opt = |\textcolor{blue}{minimize}|(Jhat, method="L-BFGS-B", tol=1.0e-7,
							   options={"disp": True, "maxiter" : 20})
\end{mintedbox}
\end{listing}

\end{document}